\documentclass[letterpaper, 10 pt, conference]{ieeeconf}
\usepackage{textcomp}
\usepackage{amsfonts,amssymb} 
\usepackage{booktabs}
\usepackage{stfloats}
\usepackage{url}
\usepackage{verbatim}
\usepackage{graphicx}
\usepackage{cite}
\usepackage[version=4]{mhchem}
\usepackage{graphicx} 
\usepackage{epsfig} 
\usepackage{mathptmx} 
\usepackage{times} 
\usepackage{amsmath} 
\usepackage{amssymb}  
\usepackage[T1]{fontenc} 
\usepackage{tabularx}  
\usepackage{hyperref}
\hypersetup{hidelinks,
	colorlinks=true,
	allcolors=black,
	pdfstartview=Fit,
	breaklinks=true}
\usepackage{multirow}
\usepackage{url}
\usepackage{makecell}
\usepackage{diagbox} 
\usepackage{bm}
\usepackage[linesnumbered,ruled,vlined]{algorithm2e}

\IEEEoverridecommandlockouts                           

\title{
 \LARGE \bf SATac: A Thermoluminescence Enabled Tactile Sensor for Concurrent Perception of Temperature, Pressure, and Shear
 }

	\author{Ziwu Song$^{*}$, Ran Yu$^{*}$, Xuan Zhang, Kit Wa Sou, Shilong Mu, Dengfeng Peng, \\Xiao-Ping Zhang,~\IEEEmembership{Fellow,~IEEE}, Wenbo Ding  
	\thanks{* The authors contributed equally to this work.}
	\thanks{This work is supported in part by Shenzhen Science and Technology Program (JCYJ20220530143013030), by Guangdong Innovative and Entrepreneurial Research Team Program (2021ZT09L197), and by Tsinghua Shenzhen International Graduate School-Shenzhen Pengrui Young Faculty Program of Shenzhen Pengrui Foundation (No. SZPR2023005).\textit{(Z.~Song and R.~Yu contributed equally to this work. Corresponding author: W.~Ding.)}}
	\thanks{Z.~Song, R.~Yu, X.~Zhang, K.~Su, X.-P.~Zhang, and W.~Ding are with Shenzhen International Graduate School, Tsinghua University, China. E-mail:\{song-zw20,yur23,x-zhang23,sujh21,msl22\}@mails.tsinghua.edu.cn, \{xiaoping.zhang,ding.wenbo\}@sz.tsinghua.edu.cn.}
    \thanks{D.~Peng is with Shenzhen University, China. E-mail:\{pengdengfeng\}@szu.edu.cn}
	\thanks{X.-P.~Zhang and W.~Ding are also with RISC-V International Open Source Laboratory, Shenzhen, China, 518055.}
	\thanks{X.-P.~Zhang is also with the Department of Electrical, Computer, and Biomedical Engineering, Ryerson University, Toronto, ON M5B 2K3, Canada.}
 }
 
\begin{document} 
\maketitle
\thispagestyle{empty}
\pagestyle{empty}

\begin{abstract}
Most vision-based tactile sensors use elastomer deformation to infer tactile information, which can not sense some modalities, like temperature. As an important part of human tactile perception, temperature sensing can help robots better interact with the environment. In this work, we propose a novel multimodal vision-based tactile sensor, SATac, which can simultaneously perceive information of temperature, pressure, and shear. SATac utilizes thermoluminescence of strontium aluminate (SA) to sense a wide range of temperatures with exceptional resolution. Additionally, the pressure and shear can also be perceived by analyzing Voronoi diagram. A series of experiments are conducted to verify the performance of our proposed sensor. We also discuss the possible application scenarios and demonstrate how SATac could benefit robot perception capabilities.

\end{abstract}

\section{Introduction}
\begin{figure}
	\centering
	\includegraphics[width=\linewidth]{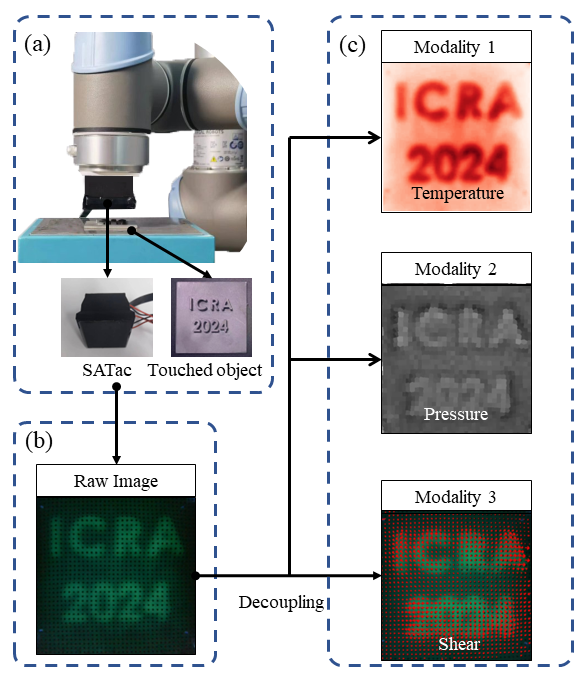}
	\caption{(a) SATac, the touched object, and the robot arm equipped with SATac exploring the environment. (b) The raw image captured by the camera inside SATac. (c) The 3 modalities including temperature, pressure, and shear decoupled from the raw image.}
	\label{fig:1}
\end{figure}

In the realm of robotic tactile perception, the pursuit of equipping robots with tactile sensing capabilities that match or even surpass human proficiency is deemed pivotal. Such advancements can significantly enhance a robot's ability to execute intricate tasks, ranging from grasping and object recognition to dexterous manipulation~\cite{park2022biomimetic, dong2021tactile, pai2023tactofind}.

\begin{figure*}
	\centering
	\includegraphics[width=1.0\linewidth]{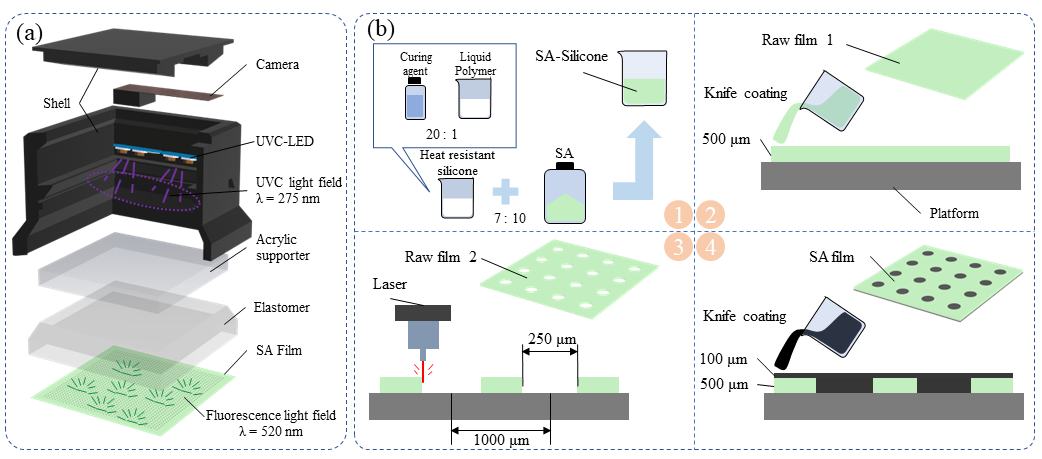}
	\caption{Design and fabrication process of SATac: (a) The exploded view of SATac. (b) The four steps to fabricate SA film: step 1, mix the heat-resistant silicone with powdered SA to make the SA-Silicone; step 2, fabricate Raw film 1 with knife coating; step 3, fabricate Raw film 2 using laser cutting on Raw film 1; step 4, fabricate the SA film with knife coating filling the black silicone in holds.}
	\label{fig:2}
\end{figure*}

Solutions for achieving robotic tactile sensing can be broadly categorized into two main types: electrical signal-based approach and vision-based approach. The former one boasts a more extensive historical foundation and is renowned for its high sensing precision. However, with the growing demand for dexterous robotic manipulation~\cite{wang2023versatile,yin2021modeling,bircher2021complex}, the significance of high-resolution spatial sensing has become evident~\cite{sun2022soft}. The vision-based approach emerges as a more competitive solution with both cost-effectiveness and robustness while ensuring real-time sampling rates, superior resolution, and an extensive sensing range~\cite{qu2023recent, abad2020visuotactile, zhang2022hardware}. These advantages have positioned it in the focal point of robotic sensing research. Vision-based tactile sensors like GelSight~\cite{johnson2009retrographic}, GelSlim~\cite{taylor2022gelslim}, DIGIT\cite{lambeta2020digit}, DTac~\cite{lin2023dtact}, etc., can reconstruct the 3D geometry of the contact object surface with high precision. Some of these, like Gelsight wedge~\cite{wang2021gelsight}, and UVTac~\cite{kim2022uvtac} can also sense the pressure and shear applied on the sensors during the contact. Besides, based on the Voronoi algorithm, the pressure, shear, and location information can be inferred simultaneously~\cite{cramphorn2018voronoi,lepora2021soft}.

Human perception is inherently multimodal; beyond deformation and force, elements such as vibration, humidity, and temperature play crucial roles in tactile sensation. Similarly, this multimodal information is vital for robotic tactile sensing~\cite{shin2022ultrasensitive, mittendorfer2011humanoid, choi2019stretchable}.
There exists some relevant research on this topic.
Authors in~\cite{park2022biomimetic} have proposed a flexible multimodal tactile sensor inspired by human skin. By utilizing both electrical impedance tomography and photoacoustic tomography techniques to interpret the electrical signals from the sensor's integrated electrode and microphone array,  information related to contact pressure and vibration can be perceived.


Temperature perception is very important for humans, for example, it can help people avoid potential harm. For robots, this capability allows them to interact better with the external environment~\cite{liang2022direct, gao2023biomimetic, ma2022bimodal}.
Authors in~\cite{shin2020sensitive} proposed a 5$\times$5 sensing array based on a Ni-NiO-Ni structure especially for temperature perception. This sensor exhibits excellent negative temperature coefficient characteristics within the range of 20 °C to 70 °C, providing high sensitivity for temperature sensing. 
There also exists research about multimodal temperature sensing. A quadruple tactile sensor has been proposed in~\cite{li2020skin}, equipped with two pairs of sensing elements arranged on either side of a layer of sponge material, which can be used to determine the pressure applied on the sensor and the temperature of the contacted object.

However, this method does not achieve high-resolution spatial perception concurrently with temperature sensing. This raises a natural question: Is it feasible to develop a tactile sensor for robotic perception that seamlessly integrates temperature sensing with high spatial resolution?

This study introduces a high-resolution, vision-based multimodal tactile sensor named SATac that can concurrently detect temperature, pressure, and shear. It utilizes Strontium Aluminate (SA)~\cite{huang2023smart,liu2023liquid}, a type of thermoluminescence material. The core sensing component of SATac is the SA film and a method for preparing SA film is also proposed. By laser-cutting small circular holes and filling them with black silicone, the SA film achieves nearly unobstructed temperature distribution, aided by guided filtering. Additionally, pressure distribution and shear force distribution during object contact are sensed by tracking changes in Voronoi polygon areas and centroid displacements. We conducted experiments on objects with complex shapes to demonstrate SATac's excellent performance in multimodal tactile perception. The main contributions of this paper can be summarized as follows:
\begin{itemize}
    \item Using SA as the functional material, we introduced SATac, a high-resolution multimodal tactile sensor capable of concurrently detecting temperature, pressure, and shear.
    \item SATac can continuously perceive multimodal data on temperature, pressure, and shear with minimal interference. This capability is achieved through the design of an Ultra Violet radiation C (UVC) light field, combined with a layered arrangement of SA film markers and complementary data processing algorithms.
    \item  The perceptual performance of SATac in its three modalities has been characterized, and the exploration and discussion of SATac's applications in different scenarios have been conducted. 
\end{itemize}

The rest of this paper is organized as follows. We present the design and fabrication of SATac in Section \ref{section 2}. We then introduce the methodology of feature extraction in Section \ref{section 3}. We evaluate the performance of SATac with a series of experiments and explore its potential application scenarios in Section \ref{section 4}. The whole paper ends by the discussion and conclusion in Section \ref{section 5}.

\begin{figure*}[t]
	\centering
	\includegraphics[width=1.0\linewidth]{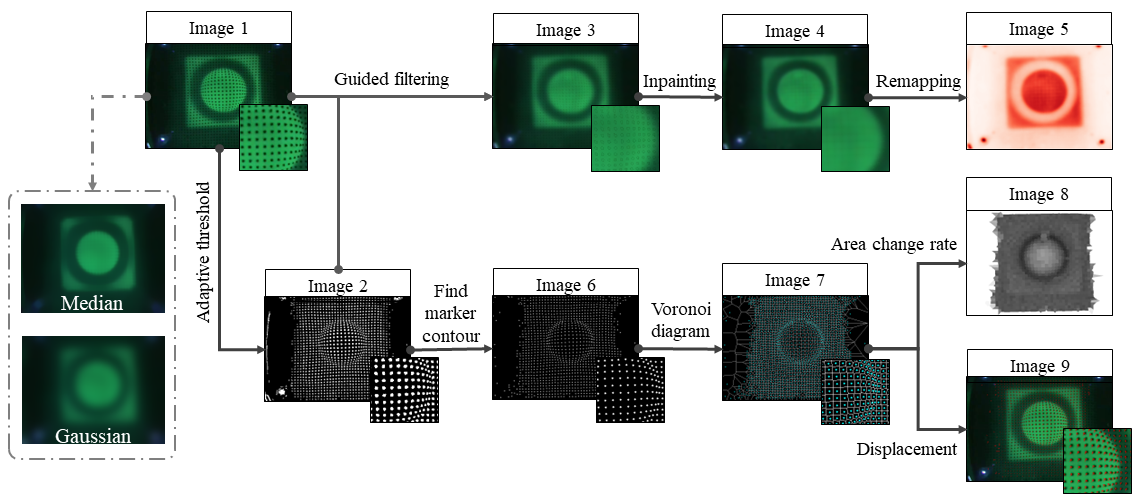}
	\caption{The complete working flow of inferring temperature (Image 5), pressure (Image 8), and shear (Image 9)  from the raw image (Image 1). Temperature information is extracted with methods including guided filtering, inpainting, and remapping. Marker contouring detection and Voronoi diagram algorithm are used for processing pressure and shear, which are calculated by area change rate and marker displacement respectively.}
	\label{fig:3}
\end{figure*}

\section{Sensor Design and Fabrication} \label{section 2}
In this section, we aim to elucidate the design and fabrication process of the SATac sensor.

\subsection{SA Material}
SA material endows SATac with the capability of perceiving temperature. As a fluorescent material with thermoluminescent properties, SA emits visible light with a wavelength near 520 nm when thermally excited, and the luminescence intensity is temperature-dependent. Typically, within the temperature range of 50 °C to 180 °C, the luminescence intensity increases as the temperature rises. Once the temperature surpasses 180 °C, the luminescence intensity gradually diminishes. With these properties, SA can be used in vision-based tactile sensing, aiding in the detection of a relatively broad range of temperature distributions.

The persistent thermoluminescence of SA necessitates external charging, which is determined by its underlying mechanism. When thermally excited, electrons in traps undergo a transition from low to high energy levels. They then revert to lower energy levels, emitting photons in the process. After releasing photons, electrons often transition to even lower energy levels than the initial state, which means that the required excitation temperature for the next cycle becomes higher. This makes thermoluminescent properties non-reusable over multiple cycles, and the relationship between temperature and luminescence intensity can change with the number of thermal excitations and the excitation temperature. In order to maintain SA at a consistent initial energy level for successive thermoluminescence events, we use ultraviolet light for charging.
This procedural step guarantees that SA consistently reverts to a stable luminosity level following each thermoluminescent event, ensuring its readiness for subsequent excitations.

\subsection{SA Film}

The SA film is made of silicone and SA powder for temperature perception. SA in its powdered form is widely used and cost-effective. When combined with silicone at a precise mass ratio, it neither disrupts the solidification process of the silicone nor compromises its elasticity. This composite, referred to as SA-Silicone, also demonstrates thermoluminescent characteristics, with luminescence intensity proportionally increasing as the temperature rises. On the SA film, there is a black marker array, which neither emits light nor transmits visible light. The diameter of each marker is 0.25 mm and the distance between two adjacent markers is 1 mm as shown in Fig.~\ref{fig:2}(a).

The whole process of SA film fabrication is illustrated in Fig.~\ref{fig:2}(b). 
First, the SA powder, characterized by a particle diameter of 100 nm, is thoroughly mixed with silicone in a mass ratio of 7:10. The mixture is subsequently placed in a negative pressure environment for 15min to eliminate air bubbles trapped within the silicone.
Second, the mixture is evenly scraped onto a flat glass substrate, forming a layer with a thickness of 500 um using knife coating. After placement in a 60 $^\text{o}$C environment for approximately 1 hour, the mixture is solidified into Raw Film 1 in step 2 of Fig.~\ref{fig:2}(b).
Third, Accurate incisions are performed on Raw film 1 using a laser emitting at a wavelength of 10400 nm, resulting in dimensions of 50$\times$50 mm$^2$ in the form of a square. Within this square region, a 40$\times$40 array of circular perforations is precisely crafted using laser cutting. Each circular aperture has a diameter of 0.25 mm, with a 1 mm spacing between adjacent apertures.
Finally, silicone mixed with black pigment, which can block visible light efficiently, is uniformly applied onto the Raw Film 2, ensuring the complete filling of the circular holes with the pigmented silicone. Upon the curing of the black silicone, the fabrication process of SA film is completed.

\subsection{Dual-light Field}
In Fig.~\ref{fig:2}(a), we present the SATac design that incorporates a dual-light field: the UVC light field and the fluorescence field. This innovative design allows the SA to revert to its original brightness post-thermoluminescence. Moreover, it equips SATac with enhanced sensing capabilities that extend beyond mere temperature perception, encompassing modalities like pressure and shear force sensing. As previously mentioned, the UVC light field, with a wavelength of 275 nm, is employed for charging the SA film. Notably, this wavelength is beyond the capture range of standard RGB cameras. On the other hand, the SA film emits a fluorescence field with a wavelength of 520 nm. This fluorescence serves as a backdrop, enabling clear detection of marker positions on the SA film.

\subsection{Overall Sensor Architecture}

SATac measures 50 mm in length, 50 mm in width, and 30 mm in height. The complete architecture of the SATac sensor is depicted in Fig.~\ref{fig:2}. Apart from the SA film that has been introducing, the remaining parts of SATac consist an outer shell, a camera, a set of UVC-LEDs, a transparent acrylic support, and a PDMS transparent elastomer. 

The outer shell is used for protection and to prevent external light from affecting the inside light fields. The camera has a field of view of 105 degrees and a frame rate of 60 FPS, with its main optical axis perpendicularly to the center of the SA film. 4 UVC-LEDs are placed around the main optical axis of the camera. Each UVC-LED operates at a voltage of 4.7 V, with a total power consumption of 0.5 W. The transparent acrylic support is used to prevent overall deformation of the elastomer, ensuring sufficient contact with objects during the touching. The PDMS elastomer, prepared by mixing liquid polymer and curing agent at a 20:1 ratio, ensures congruent deformation with the SA film. The SA film is firmly bonded to the elastomer using this PDMS mixture. The integration of a transparent elastomer with acrylic support allows the camera to distinctly capture variations in brightness and the movement of markers. Additionally, this design minimizes the obstruction of the UVC-light field, ensuring efficient charging of the SA film.

\begin{figure}
	\centering
	\includegraphics[width=1.0\linewidth]{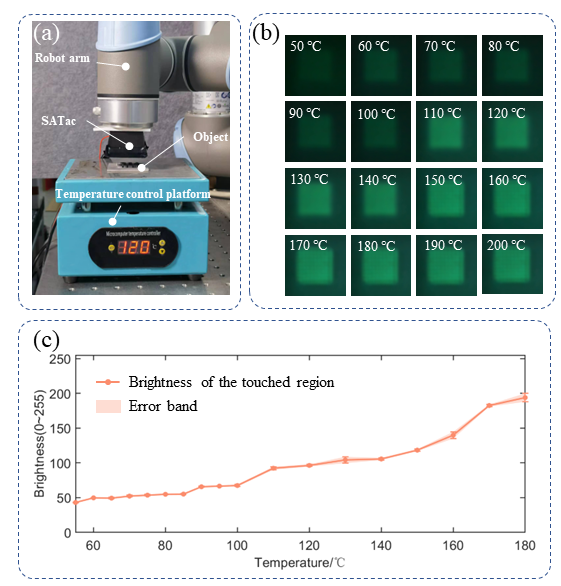}
	\caption{Characterization Experiment of Temperature Sensing. (a) Experimental setup for characterizing the performance of SATac in temperature sensing. (b) Experimental data for SATac's contact temperature when in contact with objects ranging from 50°C to 200°C. (c) Performance of SATac under different temperature conditions.}
	\label{fig:4}
\end{figure}
\section{Feature Extraction} \label{section 3}
This section introduced the details of inferring the critical features for sensing the temperature distribution, pressure distribution, and shear distribution. All the image processes are achieved using OpenCV~\cite{opencv}. 
\subsection{Temperature}

The thermoluminescence properties of SA indicate that the temperature of a contacted object influences the peak brightness of the SA film. Consequently, the brightness peak extracted from an image can depict the SA film's temperature distribution. The black marker array, being non-luminous and opaque, does not reflect temperature-induced brightness changes, leading to an incomplete temperature representation in Image 1 of Fig.~\ref{fig:3}. Image 2, derived by applying an adaptive threshold to Image 1, displays all marker contours. These contours guide the guided filtering~\cite{he2012guided,mao2022surface} of Image 1, effectively eliminating the black markers without compromising high-frequency details. Post-filtering, residual contour outlines remain, but a subsequent inpainting operation removes them, as evident in Image 4. This method preserves the SA material's temperature information and fills marker positions by referencing adjacent pixel brightness, yielding a temperature distribution free from market disruptions.

To further explain the reason for using a guided filter, we compare its filtered image with the Gaussian filter and median filter results as shown in Fig.~\ref{fig:3}. Although these two filters remove the markers successfully, excessive smoothing exists outside the markers. Comparing Image 1 and Image 4, the combination of guided filtering and impainting realizes both mitigate the influence of markers and prevent the image from distortion. This shows the effectiveness of our proposed method.

\subsection{Pressure and Shear}

The estimation of pressure and shear is achieved by analyzing the displacement of markers. In Fig.~\ref{fig:3}, Image 2 contains the contour information of the markers, after finding and detecting the marker positions, the centers of each marker are located in Image 6. With these contour centers, the Voronoi diagram of each center can be figured out, which is presented in Image 7. 

For pressure distribution, the pressure at the location of each center is presented as the area change ratio of the corresponding Voronoi diagram. In Fig.~\ref{fig:3}, Image 8 is the remapping results from the area change rate to the brightness of the gray image.  

For shear distribution, the shear at the location of each center is presented as the displacement in the image, where the displacement is defined as the difference between the position in the current frame and the position in the original frame. The tracking displacement of all the markers is visualized in Image 9 of Fig.~\ref{fig:3}, which is considered as the shear distribution. 

\section{Experiments}  \label{section 4}

\subsection{Characterization}
\subsubsection{Temperature}

In the experiment characterizing the perception of the temperature distribution, SATac is firmly mounted on the end-effector of the UR5 robot arm, as shown in Fig.~\ref{fig:4}(a). A metal cube with a 10 mm side length is placed at the center of a temperature-controlled platform. The surface temperature is controlled to increase from 50 °C to 200 °C with a 5 °C step. At each temperature value, the robot arm drives SATac to move vertically downward from a distance of 25 mm to the platform at a speed of 20 mm/s, moving a total of 16 mm. After reaching the target position, the cube is in sufficient contact with SATac, and the robot arm maintains static for 5s before returning to the starting point. Fig.~\ref{fig:4}(b) shows some selected visualization results of the temperature distribution. It can be clearly observed that the brightness gradually increases from 50 °C to 180 °C, and a decrease in brightness occurs after 180 °C. After repeating this experiment five times, we obtained the statistical results shown in Fig.~\ref{fig:4}(c). In the results of multiple trials, it is evident that the brightness steadily increases with the rise in temperature between 50 °C and 180 °C. The narrow error band indicates the stable performance of SATac in temperature perception.

\begin{figure}[htbp]
	\centering
	\includegraphics[width=1.0\linewidth]{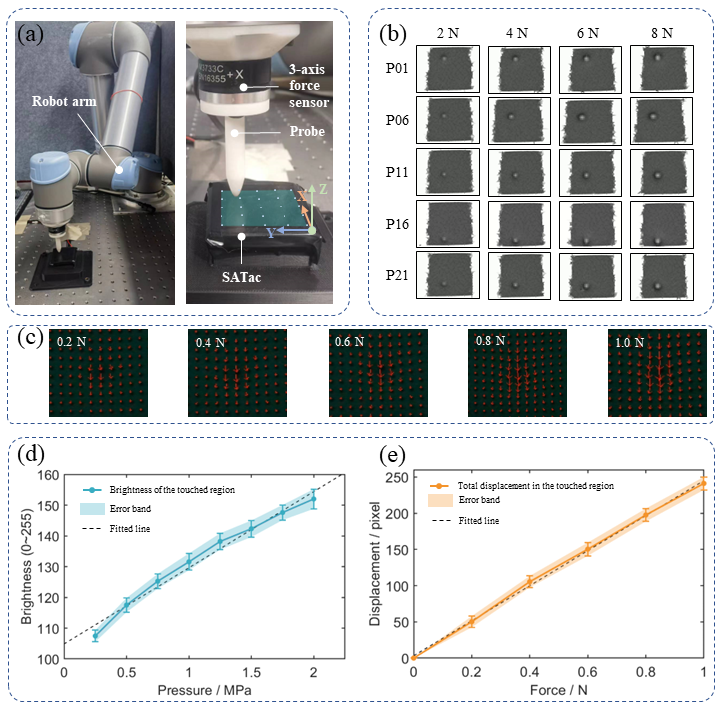}
	\caption{Characterization experiment of SATac pressure and shear sensing. (a) Experimental setup. (b) Experimental data for applying positive pressures of 2 N, 4 N, 6 N, and 8 N at pressing positions P1, P6, P11, P16, and P21. (c) Experimental data for applying shear of 0.2 N, 0.4 N, 0.6 N, 0.8 N, and 1.0 N. (d) The linear relationship between the brightness and the applied pressure. (e) The linear relationship between the marker displacement and the applied shear force.}
	\label{fig:5}
\end{figure}

\subsubsection{Pressure and Shear}
In the characterization experiment of pressure and Shear perception, the UR5 robot arm and a 6-axis force/Torque sensor (M3733C Sunrise Instruments) are used to standardize contact force and location. As shown in Fig.~\ref{fig:5}(a), SATac was firmly secured to a flat optical platform via an adapter plate to guarantee that SATac is parallel to the optical platform. A  3-axis force sensor was mounted on the robot arm, which could provide real-time force measurements along the X, Y, and Z directions shown in Fig.~\ref{fig:5}(a). A conical probe was attached to the other end of the 3-axis force sensor. Considering that the sensing range of the sensor is 45$\times$45 mm$^2$, a square region with a size of 35$\times$35 mm$^2$ is selected in the center of the SATac's sensing region, and 25 sampling points are figured out on the region in Fig.~\ref{fig:5}(a).

In the pressure characterization study, a robot arm guides a probe to apply forces from 1 N to 8 N with a step of 1 N at 25 sample points. The probe initially positioned 20 mm above the sensor center, descends at 5 mm/s toward the target. For each force increment (from 1 N to 8 N), once the 6-axis force sensor indicates the desired force, the arm halts, maintaining pressure for 5s, then continues descending. This is repeated until an 8 N force is sustained for 5s. Subsequently, the probe returns to the 1 N position and moves in the X direction. When the force output in the X direction hits increments from 0.2 N to 1.0 N, the same 5s pause procedure is implemented. The process concludes by returning the probe to its initial position, and this cycle is performed across all 25 sampling points.

\begin{figure}[t]
	\centering
	\includegraphics[width=1.0\linewidth]{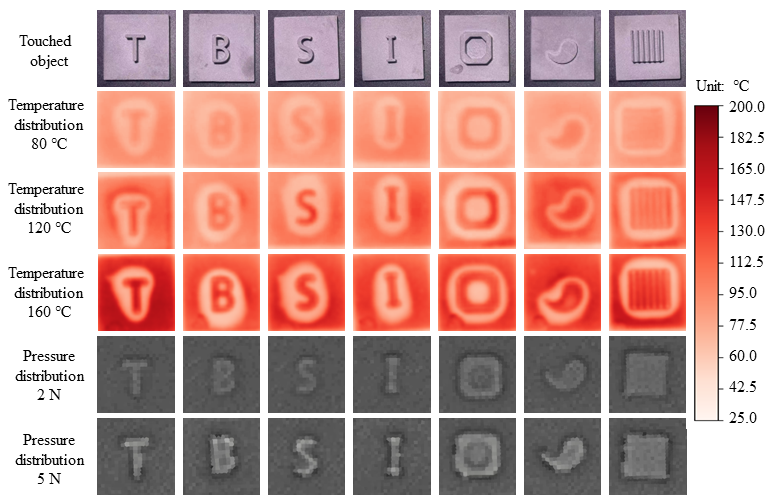}
	\caption{SATac touches different objects for testing their properties, and obtains corresponding temperature distribution (80 °C, 120 °C, and 160 °C) and pressure distribution (2 N and 5 N).}
	\label{fig:6}
\end{figure}

Some selected pressure distributions output by SATac are shown in Fig.~\ref{fig:5}(b). It can be observed from the result that greater forces are exerted at varying positions, while increased brightness coincides with these locations in the remapped pressure distribution. The relationship between pressure and brightness is presented in Fig.~\ref{fig:5}(c). Fig.~\ref{fig:5}(c) shows some selected shear distributions from SATac. The relationship between shear and the cumulative displacement of all markers is presented in Fig.~\ref{fig:5}(d). For both pressure and shear experiments, we conduct multiple repeated experiments to calculate the corresponding error bars and error band. The results reveal a strong linear correlation between brightness and both pressure and shear, with relatively small errors.

\subsection{Application}
\begin{figure*}[htbp]
	\centering
	\includegraphics[width=1.0\linewidth]{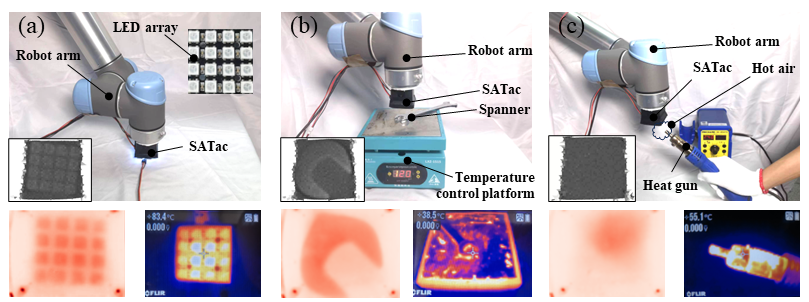}
	\caption{Performance of SATac in Four Scenarios. (a) SATac touches a working LED array circuit board. (b) SATac touches a metallic object with a high temperature. (c) In a non-contact situation, SATac senses a high-temperature gas atmosphere.}
	\label{fig:7}
\end{figure*}

SATac possesses the capability to concurrently acquire multimodal information with exceptional resolution. Experiments were conducted by affixing SATac to the end of a robotic arm, enabling it to sense objects of varying complex shapes. Fig.~\ref{fig:6} illustrates the perception results under varying temperature and contact pressure conditions. These results demonstrates commendable ability in acquiring multimodal information, as evidenced by its robust performance in diverse experimental conditions.

In certain scenarios, the commonly-used high-resolution temperature measurement method can not accurately measure the temperature of target objects, and SATac offers a solution to this challenge.
Fig.~\ref{fig:7}(a), we simulate one possible scenario where an illuminated LED array with each LED is at 80 °C. The thermal imager reveals a central LED cluster with a higher temperature than its periphery, which is not in line with the actual situation. This is caused by interference between different lights. Similarly, in Fig.~\ref{fig:7}(b) and Fig.~\ref{fig:7}(c), the thermal imager struggles to measure the temperature accurately due to metal reflection and the passage of infrared light through hot air, respectively. SATac shows better performances when dealing with those situations. Additionally, sensing results of SATac in Fig.~\ref{fig:7} (a) and (b) also show multimodal ability by measuring pressure distribution. Considering the result of Fig.~\ref{fig:7} (c), although there is no remarkable deformation on the sensor's surface, multimodal ability enables SATac to sense the hot air, which can be hardly detected by conventional vision-based sensors.

\begin{table}[htbp]
	\caption{Comparison of the vision-based tactile sensors}
	\centering
        \resizebox{\linewidth}{!}{
	\begin{tabular}{c|cccc}
		\hline \hline
		\ & Size (mm) & \makecell{Sensing Field \\(mm$^{2}$)} & Resolution & Modality\\
		\hline
		DIGIT\cite{lambeta2020digit} & 20$\times$27$\times$18 & 19$\times$16 & 640$\times$480 & 3D deformation\\
       
		DTact\cite{lin2023dtact} &  45$\times$45$\times$47  & 24$\times$24 & 800$\times$600 & 3D deformation\\

        DotView\cite{zheng2023dotview} & 25.5$\times$47$\times$14.4 & 8×8  & 192$\times$192 & pressure and shear\\

		\textbf{DelTact\cite{zhang2022deltact}} & 39$\times$60$\times$30 & 36$\times$34 & 1280$\times$720 & 3D deformation, pressure and shear\\

		TacRot\cite{zhang2022tacrot} & 42$\times$84$\times$45 & N/A  & 800$\times$600 & 3D deformation, pressure and shear\\

		\textbf{SATac (Ours)} & \textbf{50$\times$50$\times$30} & \textbf{40$\times$40} & \textbf{640$\times$480} & \textbf{temperature, pressure and shear} \\
		\hline
	\end{tabular}}
	\label{tab1}
\end{table}

\section{Discussions and Conclusion} \label{section 5}
This paper introduces SATac, a high-resolution multimodal vision-based tactile sensor based on the thermoluminescence material. SATac is capable of simultaneously perceiving temperature distribution, pressure distribution, and shear force distribution of the contacted object. Moreover, due to its high resolution, it can also discern object shapes and spatial positions of contact areas. The core sensing component of SATac is the SA film and a method for preparing SA film is proposed. By laser-cutting small circular holes and filling them with black silicone, the SA film achieves nearly unobstructed temperature distribution, aided by guided filtering. Additionally, pressure distribution and shear force distribution during object contact are sensed by tracking changes in Voronoi polygon areas and centroid displacements. We conducted experiments on objects with complex shapes to demonstrate SATac's excellent performance in multimodal tactile perception.

Compared to previous classical multimodal tactile sensing approaches~\ref{tab1}, SATac offers the advantage of perceiving temperature, pressure, and shear force simultaneously while also providing information about object shape and contact location. It encompasses a broader range of sensing modalities and boasts a considerable sensing area and sampling frequency. 

However, as described in this paper, SATac's temperature sensing range falls between 50 $^\text{o}$C to 180 $^\text{o}$C, which is within the high-temperature range for everyday scenarios. In the future, we can explore the possibility of customizing the temperature sensing range by introducing specific ion doping to the SA material, allowing it to cover a broader temperature range.

%

\bibliographystyle{IEEEtran}
\bibliography{ref}
\end{document}